\documentclass[conference]{IEEEtran}
\IEEEoverridecommandlockouts
\usepackage{cite}
\usepackage{tikz}
\usepackage{algorithm}
\usepackage{algpseudocode}
\usetikzlibrary{shapes,arrows}
\tikzstyle{decision} = [diamond, draw, fill=blue!20, 
    text width=4.5em, text badly centered, node distance=3cm, inner sep=0pt]
\tikzstyle{block} = [rectangle, draw, fill=blue!20, 
    text width=5em, text centered, rounded corners, minimum height=4em]
\tikzstyle{line} = [draw, -latex']
\tikzstyle{cloud} = [draw, ellipse,fill=red!20, node distance=3cm,
    minimum height=2em]
\usepackage{amsmath,amssymb,amsfonts}

\usepackage{graphicx}
\usepackage{textcomp}
\usepackage{epic,epsfig}
\usepackage{xcolor}
\def\BibTeX{{\rm B\kern-.05em{\sc i\kern-.025em b}\kern-.08em
    T\kern-.1667em\lower.7ex\hbox{E}\kern-.125emX}}
\begin{document}

%
%

\title{LAXARY: A Trustworthy Explainable Twitter Analysis Model for Post-Traumatic Stress Disorder Assessment\\
}

\author{\IEEEauthorblockN{ \textsubscript{1}Mohammad Arif Ul Alam, \textsubscript{2}Dhawal Kapadia}
\IEEEauthorblockA{\textit{\textsuperscript{1}Department of Computer Science, University of Massachusetts Lowell}\\
\textit{\textsuperscript{2}IQVIA, Manhattan, Newyork}\\
mohammadariful\_alam@uml.edu,dhawalkapadia8@gmail.com}
}

\maketitle

\begin{abstract}
Veteran mental health is a significant national problem as large number of veterans are returning from the recent war in Iraq and continued military presence in Afghanistan. While significant existing works have investigated twitter posts-based Post Traumatic Stress Disorder (PTSD) assessment using blackbox machine learning techniques, these frameworks cannot be trusted by the clinicians due to the lack of clinical explainability. To obtain the trust of clinicians, we explore the big question, can twitter posts provide enough information to fill up clinical PTSD assessment surveys that have been traditionally trusted by clinicians? To answer the above question, we propose, LAXARY (Linguistic Analysis-based Exaplainable Inquiry) model, a novel Explainable Artificial Intelligent (XAI) model to detect and represent PTSD assessment of twitter users using a modified Linguistic Inquiry and Word Count (LIWC) analysis. First, we employ clinically validated survey tools for collecting clinical PTSD assessment data from real twitter users and develop a PTSD Linguistic Dictionary using the PTSD assessment survey results. Then, we use the PTSD Linguistic Dictionary along with machine learning model to fill up the survey tools towards detecting PTSD status and its intensity of corresponding twitter users. Our experimental evaluation on 210 clinically validated veteran twitter users provides promising accuracies of both PTSD classification and its intensity estimation. We also evaluate our developed PTSD Linguistic Dictionary's reliability and validity.
\end{abstract}



\section{Introduction}
Combat veterans diagnosed with PTSD are substantially more likely to engage in a number of high risk activities including engaging in interpersonal violence, attempting suicide, committing suicide, binge drinking, and drug abuse \cite{rosen05}. Despite improved diagnostic screening, outpatient mental health and inpatient treatment for PTSD, the syndrome remains treatment resistant, is typically chronic, and is associated with numerous negative health effects and higher treatment costs \cite{rosenheck99}. As a result, the Veteran Administration's National Center for PTSD (NCPTSD) suggests to reconceptualize PTSD not just in terms of a psychiatric symptom cluster, but focusing instead on the specific high risk behaviors associated with it, as these may be directly addressed though behavioral change efforts \cite{rosen05}. Consensus prevalence estimates suggest that PTSD impacts between 15-20\% of the veteran population which is typically chronic and treatment resistant \cite{rosen05}. The PTSD patients support programs organized by different veterans peer support organization use a set of surveys for local weekly assessment to detect the intensity of PTSD among the returning veterans. However, recent advanced evidence-based care for PTSD sufferers surveys have showed that veterans, suffered with chronic PTSD are reluctant in participating assessments to the professionals which is another significant symptom of war returning veterans with PTSD. Several existing researches showed that, twitter posts of war veterans could be a significant indicator of their mental health and could be utilized to predict PTSD sufferers in time before going out of control \cite{rude04,ram08,d11,alvarez01,kramer04,chung07,park12}. However, all of the proposed methods relied on either blackbox machine learning methods or language models based sentiments extraction of posted texts which failed to obtain acceptability and trust of clinicians due to the lack of their explainability.

In the context of the above research problem, we aim to answer the following {\bf research questions}

\begin{itemize}
    \item Given clinicians have trust on clinically validated PTSD assessment surveys, can we fill out PTSD assessment surveys using twitter posts analysis of war-veterans?
    \item If possible, what sort of analysis and approach are needed to develop such XAI model to detect the prevalence and intensity of PTSD among war-veterans only using the social media (twitter) analysis where users are free to share their everyday mental and social conditions?
    \item How much quantitative improvement do we observe in our model's ability to explain both detection and intensity estimation of PTSD?
    
\end{itemize}
In this paper, we propose LAXARY, an explainable and trustworthy representation of PTSD classification and its intensity for clinicians.
 
The {\bf key contributions} of our work are summarized below,
\begin{itemize}
\item The novelty of LAXARY lies on the proposed clinical surveys-based PTSD Linguistic dictionary creation with words/aspects which represents the instantaneous perturbation of twitter-based sentiments as a specific pattern and help calculate the possible scores of each survey question.

\item LAXARY includes a modified LIWC model to calculate the possible scores of each survey question using PTSD Linguistic Dictionary to fill out the PTSD assessment surveys which provides a practical way not only to determine fine-grained discrimination of physiological and psychological health markers of PTSD without incurring the expensive and laborious in-situ laboratory testing or surveys, but also obtain trusts of clinicians who are expected to see traditional survey results of the PTSD assessment.

\item Finally, we evaluate the accuracy of LAXARY model performance and reliability-validity of generated PTSD Linguistic Dictionary using real twitter users' posts. Our results show that, given normal weekly messages posted in twitter, LAXARY can provide very high accuracy in filling up surveys towards identifying PTSD ($\approx 89\%$) and its intensity ($\approx 1.8$ mean squared error).

\end{itemize}

\section{Overview}
Our overall framework consists of the following logical steps: (i) Develop PTSD Detection System using twitter posts of war-veterans(ii) design real surveys from the popular symptoms based mental disease assessment surveys; (iii) define single category and create PTSD Linguistic Dictionary for each survey question and multiple aspect/words for each question; (iv) calculate $\alpha$-scores for each category and dimension based on linguistic inquiry and word count as well as the aspects/words based dictionary; (v) rank features according to the contributions of achieving separation among categories associated with different $\alpha$-scores; and select feature sets that minimize the overlap among categories as associated with the target classifier (SGD); and finally (vi) estimate the quality of selected features-based classification for filling up surveys based on classified categories i.e. PTSD assessment which is trustworthy among the psychiatry community.

\section{Related Works}
Twitter activity based mental health assessment has been utmost importance to the Natural Language Processing (NLP) researchers and social media analysts for decades. Several studies have turned to social media data to study mental health, since it provides an unbiased collection of a person's language and behavior, which has been shown to be useful in diagnosing conditions. \cite{bergsma12} used n-gram language model (CLM) based s-score measure setting up some user centric emotional word sets. \cite{chowd13} used  positive and negative PTSD data to train three classifiers: (i) one unigram language model (ULM); (ii) one character n-gram language model (CLM); and 3) one from the LIWC categories $\alpha$-scores and found that last one gives more accuracy than other ones. \cite{cop14} used two types of $s$-scores taking the ratio of negative and positive language models. Differences in language use have been observed in the personal writing of students who score highly on depression scales \cite{rude04}, forum posts for depression \cite{ram08}, self narratives for PTSD (\cite{d11,alvarez01}), and chat rooms for bipolar \cite{kramer04}. Specifically in social media, differences have previously been observed between depressed and control groups (as assessed by internet-administered batteries) via LIWC: depressed users more frequently use first person pronouns (\cite{chung07}) and more frequently use negative emotion words and anger words on Twitter, but show no differences in positive emotion word usage (\cite{park12}). Similarly, an increase in negative emotion and first person pronouns, and a decrease in third person pronouns, (via LIWC) is observed, as well as many manifestations of literature findings in the pattern of life of depressed users (e.g., social engagement, demographics) (\cite{chowd13b}). Differences in language use in social media via LIWC have also been observed between PTSD and control groups (\cite{cop14b}).

All of the prior works used some random dictionary related to the human sentiment (positive/negative) word sets as category words to estimate the mental health but very few of them addressed the problem of explainability of their solution to obtain trust of clinicians. Islam et. al proposed an explainable topic modeling framework to rank different mental health features using Local Interpretable Model-Agnostic Explanations and visualize them to understand the features involved in mental health status classification using the  \cite{tunaz19} which fails to provide trust of clinicians due to its lack of interpretability in clinical terms. In this paper, we develop LAXARY model where first we start investigating clinically validated survey tools which are trustworthy methods of PTSD assessment among clinicians, build our category sets based on the survey questions and use these as dictionary words in terms of first person singular number pronouns aspect for next level LIWC algorithm. Finally, we develop a modified LIWC algorithm to estimate survey scores (similar to sentiment category scores of naive LIWC) which is both explainable and trustworthy to clinicians.

\section{Demographics of Clinically Validated PTSD Assessment Tools}
There are many clinically validated PTSD assessment tools that are being used both to detect the prevalence of PTSD and its intensity among sufferers. Among all of the tools, the most popular and well accepted one is Domain-Specific Risk-Taking (DOSPERT) Scale \cite{dospert06}. This is a psychometric scale that assesses risk taking in five content domains: financial decisions (separately for investing versus gambling), health/safety, recreational, ethical, and social decisions. Respondents rate the likelihood that they would engage in domain-specific risky activities (Part I). An optional Part II assesses respondents' perceptions of the magnitude of the risks and expected benefits of the activities judged in Part I. There are more scales that are used in risky behavior analysis of individual's daily activities such as, The Berlin Social Support Scales (BSSS) \cite{bsss} and Values In Action Scale (VIAS) \cite{vias}. Dryhootch America \cite{dryhootch,rizia15}, a veteran peer support community organization, chooses 5, 6 and 5 questions respectively from the above mentioned survey systems to assess the PTSD among war veterans and consider rest of them as irrelevant to PTSD. The details of dryhootch chosen survey scale are stated in Table~\ref{tab:dryhootch_scale}. Table!\ref{tab:sample_scale} shows a sample DOSPERT scale demographic chosen by dryhootch. The threshold (in Table~\ref{tab:dryhootch_scale}) is used to calculate the risky behavior limits. For example, if one individual's weekly DOSPERT score goes over 28, he is in critical situation in terms of risk taking symptoms of PTSD. Dryhootch defines the intensity of PTSD into four categories based on the weekly survey results of all three clinical survey tools (DOSPERT, BSSS and VIAS )

\begin{itemize}
\item {\it High risk PTSD}:  If one individual veteran's weekly PTSD assessment scores go above the threshold for all three PTSD assessment tools i.e. DOSPERT, BSSS and VIAS, then he/she is in high risk situation which needs immediate mental support to avoid catastrophic effect of individual's health or surrounding people's life.

\item {\it Moderate risk PTSD}:  If one individual veteran's weekly PTSD assessment scores go above the threshold for any two of the three PTSD assessment tools, then he/she is in moderate risk situation which needs close observation and peer mentoring to avoid their risk progression.

\item {\it Low risk PTSD}:   If one individual veteran's weekly PTSD assessment scores go above the threshold for any one of the three PTSD assessment tools, then he/she has light symptoms of PTSD.

\item {\it No PTSD}: If one individual veteran's weekly PTSD assessment scores go below the threshold for all three PTSD assessment tools, then he/she has no PTSD.

\end{itemize}

\begin{table}[h]

    \centering
    \begin{tabular}{|l l l l|}
	\hline
	Tool & D & B & V \\
	\hline
      questions & 8 & 3 &5\\
      \hline
      chosen & 5 & 6 &5\\
      \hline
      total points &  35 & 18 &25\\
      \hline
      threshold & 28 & 13 &15\\
      \hline
    \end{tabular}
    \caption{Dryhootch chosen PTSD assessment surveys (D: DOSPERT, B: BSSS and V: VIAS) demographics}~\label{tab:dryhootch_scale}
\end{table}

\begin{table}[h]
    \centering
    \begin{tabular}{|p{8cm}|}
    \hline
      \textbf{Questions}: \textbf{1:} Betting a day's income at the horse races; \textbf{2:} Drinking heavily at a social function; \textbf{3:} Disagreeing with an authority figure on a major issue; \textbf{4:} Engaging in unprotected sex; \textbf{5:} Leaving your young children alone at home while running an errand\\
      \hline
      \textbf{Answers (Scores)}: \textbf{1:} Extremely Unlikely; \textbf{2:} Moderately Unlikely; \textbf{3:} Somewhat Unlikely; \textbf{4:} Not Sure; \textbf{5:} Somewhat Likely; \textbf{6:} Moderately Likely; \textbf{7:} Extremely Likely; and \textbf{0:} Skip Question\\
      \hline
      
      \end{tabular}
    \caption{Sample Dryhootch chosen questions from DOSPERT}~\label{tab:sample_scale}

\end{table}

\section{Twitter-based PTSD Detection}
To develop an explainable model, we first need to develop twitter-based PTSD detection algorithm. In this section, we describe the data collection and the development of our core LAXARY model.

\subsection{Data Collection}
We use an automated regular expression based searching to find potential veterans with PTSD in twitter, and then refine the list manually. First, we select different keywords to search twitter users of different categories. For example, to search self-claimed diagnosed PTSD sufferers, we select keywords related to PTSD for example, post trauma, post traumatic disorder, PTSD etc. We use a regular expression to search for statements where the user self-identifies as being diagnosed with PTSD. For example, Table~\ref{tab:sample_tweet} shows a self-identified tweet posts. To search veterans, we mostly visit to different twitter accounts of veterans organizations such as "MA Women Veterans @WomenVeterans", "Illinois Veterans @ILVetsAffairs", "Veterans Benefits @VAVetBenefits" etc. We define an inclusion criteria as follows: \emph{\bf one twitter user will be part of this study if he/she describes himself/herself as a veteran in the introduction and have at least 25 tweets in last week}. After choosing the initial twitter users, we search for self-identified PTSD sufferers who claim to be diagnosed with PTSD in their twitter posts. We find 685 matching tweets which are manually reviewed to determine if they indicate a genuine statement of a diagnosis for PTSD. Next, we select the username that authored each of these tweets and retrieve last week's tweets via the Twitter API. We then filtered out users with less than 25 tweets and those whose tweets were not at least 75\% in English (measured using an automated language ID system.) This filtering left us with 305 users as positive examples. We repeated this process for a group of randomly selected users. We randomly selected 3,000 twitter users who are veterans as per their introduction and have at least 25 tweets in last one week. After filtering (as above) in total 2,423 users remain, whose tweets are used as negative examples developing a 2,728 user's entire weeks' twitter posts where 305 users are self-claimed PTSD sufferers. We distributed Dryhootch chosen surveys among 1,200 users (305 users are self claimed PTSD sufferers and rest of them are randomly chosen from previous 2,423 users) and received 210 successful responses. Among these responses, 92 users were diagnosed as PTSD by any of the three surveys and rest of the 118 users are diagnosed with NO PTSD. Among the clinically diagnosed PTSD sufferers, 17 of them were not self-identified before. However, 7 of the self-identified PTSD sufferers are assessed with no PTSD by PTSD assessment tools. The response rates of PTSD and NO PTSD users are 27\% and 12\%. In summary, we have collected one week of tweets from 2,728 veterans where 305 users claimed to have diagnosed with PTSD. After distributing Dryhootch surveys, we have a dataset of 210 veteran twitter users among them 92 users are assessed with PTSD and 118 users are diagnosed with no PTSD using clinically validated surveys. The severity of the PTSD are estimated as Non-existent, light, moderate and high PTSD based on how many surveys support the existence of PTSD among the participants according to dryhootch manual \cite{dryhootch,rizia15}.

\begin{figure}
 \centering
  \begin{center}

  \includegraphics[trim = 0mm 0mm 0mm 0mm, clip, scale=0.4]{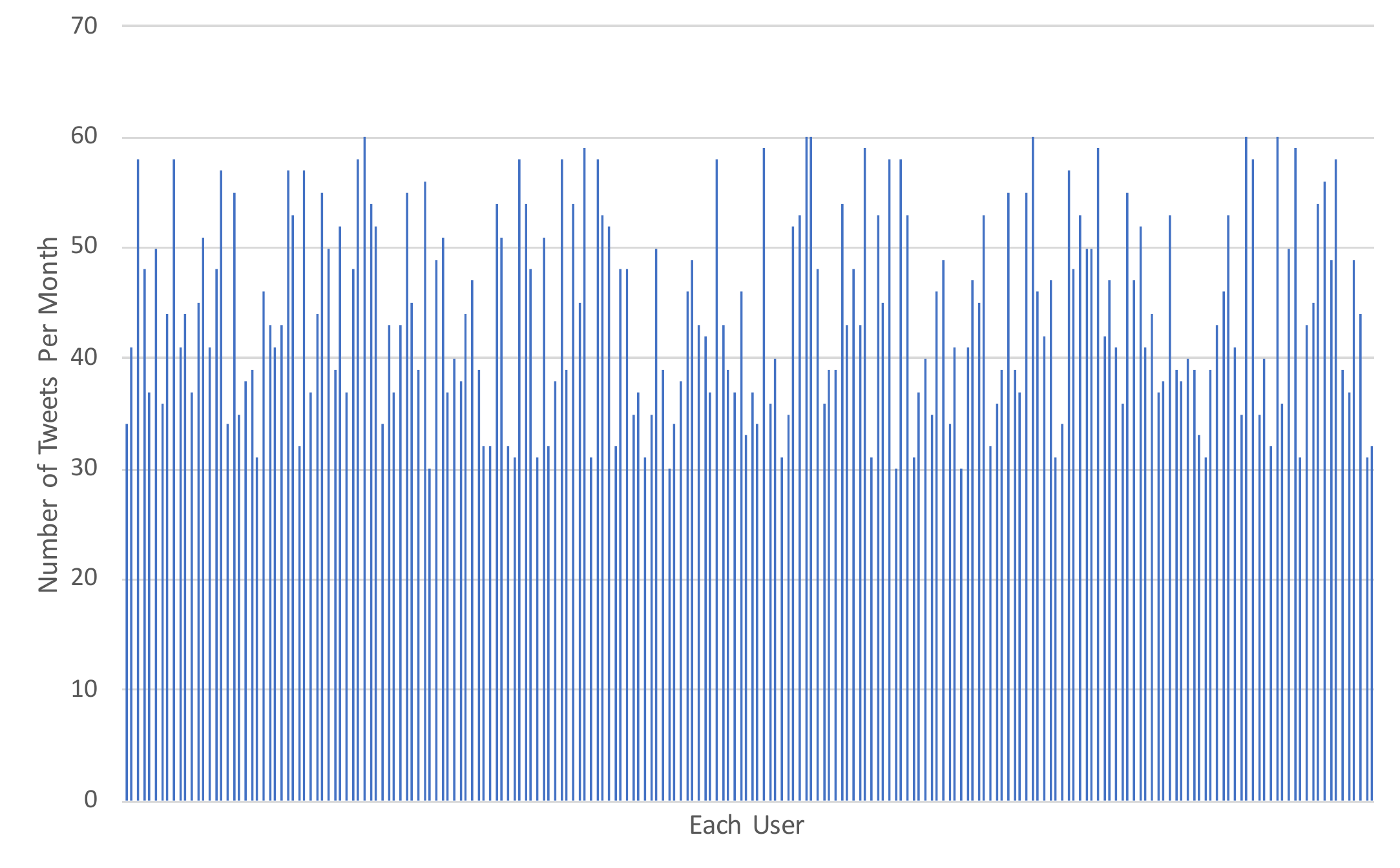}
  \caption{Each 210 users' average tweets per month}
  \label{fig:users_tweets}
  \end{center}
\end{figure}

\subsection{Pre-processing}
We download 210 users' all twitter posts who are war veterans and clinically diagnosed with PTSD sufferers as well which resulted a total 12,385 tweets. Fig~\ref{fig:users_tweets} shows each of the 210 veteran twitter users' monthly average tweets. We categorize these Tweets into two groups: Tweets related to work and Tweets not related to work. That is, only the Tweets that use a form of the word “work*” (e.g. work,worked, working, worker, etc.) or “job*” (e.g. job, jobs, jobless, etc.) are identified as work-related Tweets, with the remaining categorized as non-work-related Tweets. This categorization method increases the likelihood that most Tweets in the work group are indeed talking about work or job; for instance, “Back to work. Projects are firing back up and moving ahead now that baseball is done.” This categorization results in 456 work-related Tweets, about 5.4\% of all Tweets written in English (and 75 unique Twitter users). To conduct weekly-level analysis, we consider three categorizations of Tweets (i.e. overall Tweets, work-related Tweets, and non work-related Tweets) on a daily basis, and create a text file for each week for each group.

\subsection{PTSD Detection Baseline Model}
We use Coppersmith proposed PTSD classification algorithm to develop our baseline blackbox model \cite{cop14}. We utilize our positive and negative PTSD data (+92,-118) to train three classifiers: (i) unigram language model (ULM) examining individual whole words, (ii) character n-gram language model (CLM), and (iii)  LIWC based categorical models above all of the prior ones. The LMs have been shown effective for Twitter classification tasks \cite{bergsma12} and LIWC has been previously used for analysis of mental health in Twitter \cite{chowd13}. The language models measure the probability that a word (ULM) or a string of characters (CLM) was generated by the same underlying process as the training data. We first train one of each language model ($clm^{+}$ and $ulm^{+}$) from the tweets of PTSD users, and  another model ($clm^{-}$ and $ulm^{-}$) from the tweets from No PTSD users. Each test tweet $t$ is scored by comparing probabilities from each LM called $\alpha-score =  \frac{lm^{+}(t)}{lm^{-}(t)} $. A threshold of 1 for $\alpha-score$ divides scores into positive and negative classes. In a multi-class setting, the algorithm minimizes the cross entropy, selecting the model with the highest probability. For each user, we calculate the proportion of tweets scored positively by each LIWC category. These proportions are used as a feature vector in a loglinear regression model \cite{Pedregosa11}. Prior to training, we preprocess the text of each tweet: we replace all usernames with a single token (USER), lowercase all text, and remove extraneous whitespace. We also exclude any tweet that contained a URL, as these often pertain to events external to the user.

We conduct a LIWC analysis of the PTSD and non-PTSD tweets to determine if there are differences in the language usage of PTSD users. We applied the LIWC battery and examined the distribution of words in their language. Each tweet was tokenized by separating on whitespace. For each user, for a subset of the LIWC categories, we measured the proportion of tweets that contained at least one word from that category. Specifically, we examined the following nine categories: first, second and third person pronouns, swear, anger, positive emotion, negative emotion, death, and anxiety words. Second person pronouns were used significantly less often by PTSD users, while third person pronouns and words about anxiety were used significantly more often.

\section{LAXARY: Explainable PTSD Detection Model}
The heart of LAXARY framework is the construction of PTSD Linguistic Dictionary. Prior works show that linguistic dictionary based text analysis has been much effective in twitter based sentiment analysis \cite{amanda14,lee19}. Our work is the first of its kind that develops its own linguistic dictionary to explain automatic PTSD assessment to confirm trustworthiness to clinicians.

\begin{figure}

 \centering
  \begin{center}
  \includegraphics[trim = 0mm 0mm 0mm 0mm, clip, scale=0.4]{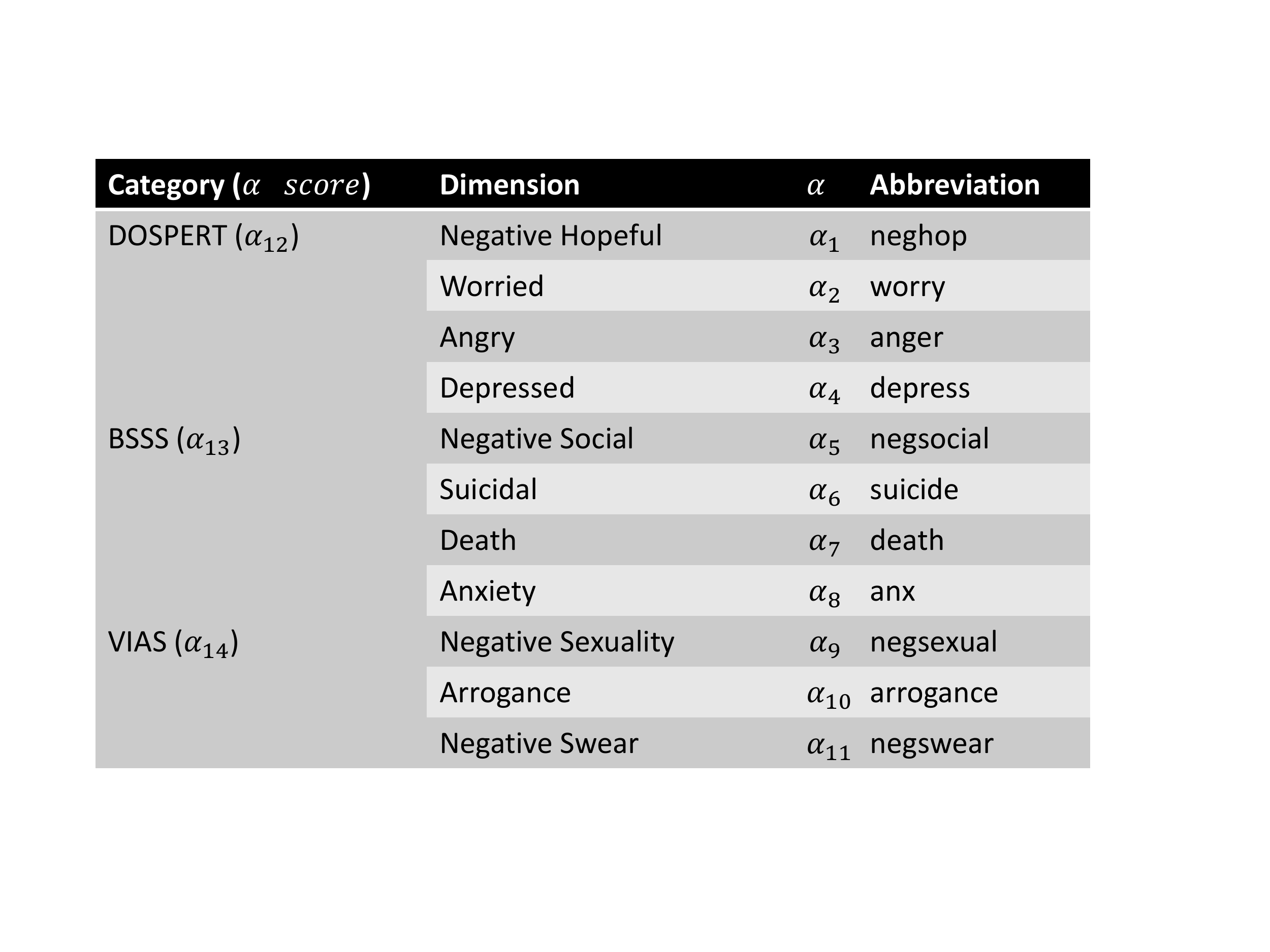}
  \caption{Category Details}
  \label{fig:category}
  \end{center}
\end{figure}

\subsection{PTSD Linguistic Dictionary Creation}
We use LIWC developed WordStat dictionary format for our text analysis \cite{liwc}. The LIWC application relies on an internal default dictionary that defines which words should be counted in the target text files. To avoid confusion in the subsequent discussion, text words that are read and analyzed by WordStat are referred to as target words. Words in the WordStat dictionary file will be referred to as dictionary words. Groups of dictionary words that tap a particular domain (e.g., negative emotion words) are variously referred to as subdictionaries or word categories. There are several steps to use this dictionary which are stated as follows:\\
      \textbf{Pronoun selection:} At first we have to define the pronouns of the target sentiment. Here we used first person singular number pronouns (i.e., I, me, mine etc.) that means we only count those sentences or segments which are only related to first person singular number i.e., related to the person himself. \\
      \textbf{Category selection:} We have to define the categories of each word set thus we can analyze the categories as well as dimensions' text analysis scores. We chose three categories based on the three different surveys: 1) DOSPERT scale; 2) BSSS scale; and 3) VIAS scale. \\
      \textbf{Dimension selection:} We have to define the word sets (also called dimension) for each category. We chose one dimension for each of the questions under each category to reflect real survey system evaluation. Our chosen categories are state in Fig~\ref{fig:category}.\\
       \textbf{Score calculation} $\alpha$-score: $\alpha$-scores refer to the Cronbach's alphas for the internal reliability of the specific words within each category. The binary alphas are computed on the ratio of occurrence and non-occurrence of each dictionary word whereas the raw or uncorrected alphas are based on the percentage of use of each of the category words within texts.

\subsection{Feature Extraction and Survey Score Estimation}
We use the exact similar method of LIWC to extract $\alpha$-scores for each dimension and categories except we use our generated PTSD Linguistic Dictionary for the task \cite{liwc}. Thus we have total 16 $\alpha$-scores in total. Then, the final output is a 16 valued matrix which represent the score for each questions from three different Dryhootch surveys. We use the output to fill up each survey, estimate the prevalence of PTSD and its intensity based on each tool's respective evaluation metric.

\section{Experimental Evaluation}
To validate the performance of LAXARY framework, we first divide the entire 210 users' twitter posts into training and test dataset. Then, we first developed PTSD Linguistic Dictionary from the twitter posts from training dataset and apply LAXARY framework on test dataset.

\begin{table}[h]

    \centering
    \begin{tabular}{|p{8cm}|}
    \hline
\emph{In loving memory my mom, she was only 42, I was 17 and taken away from me. \emph{ I was diagnosed with having P.T.S.D} LINK So today I started therapy, she diagnosed me with anorexia, depression, anxiety disorder, post traumatic stress disorder and wants me to @USERNAME The VA diagnosed me with PTSD, so I can’t go in that direction anymore I wanted to share some things that have been helping me heal lately. I was diagnosed with severe complex PTSD and... LINK}\\
\hline
      \end{tabular}
    \caption{Example of PTSD user's twitter post}~\label{tab:sample_tweet}
\end{table}

\begin{figure}
\begin{center}
 \epsfig{file=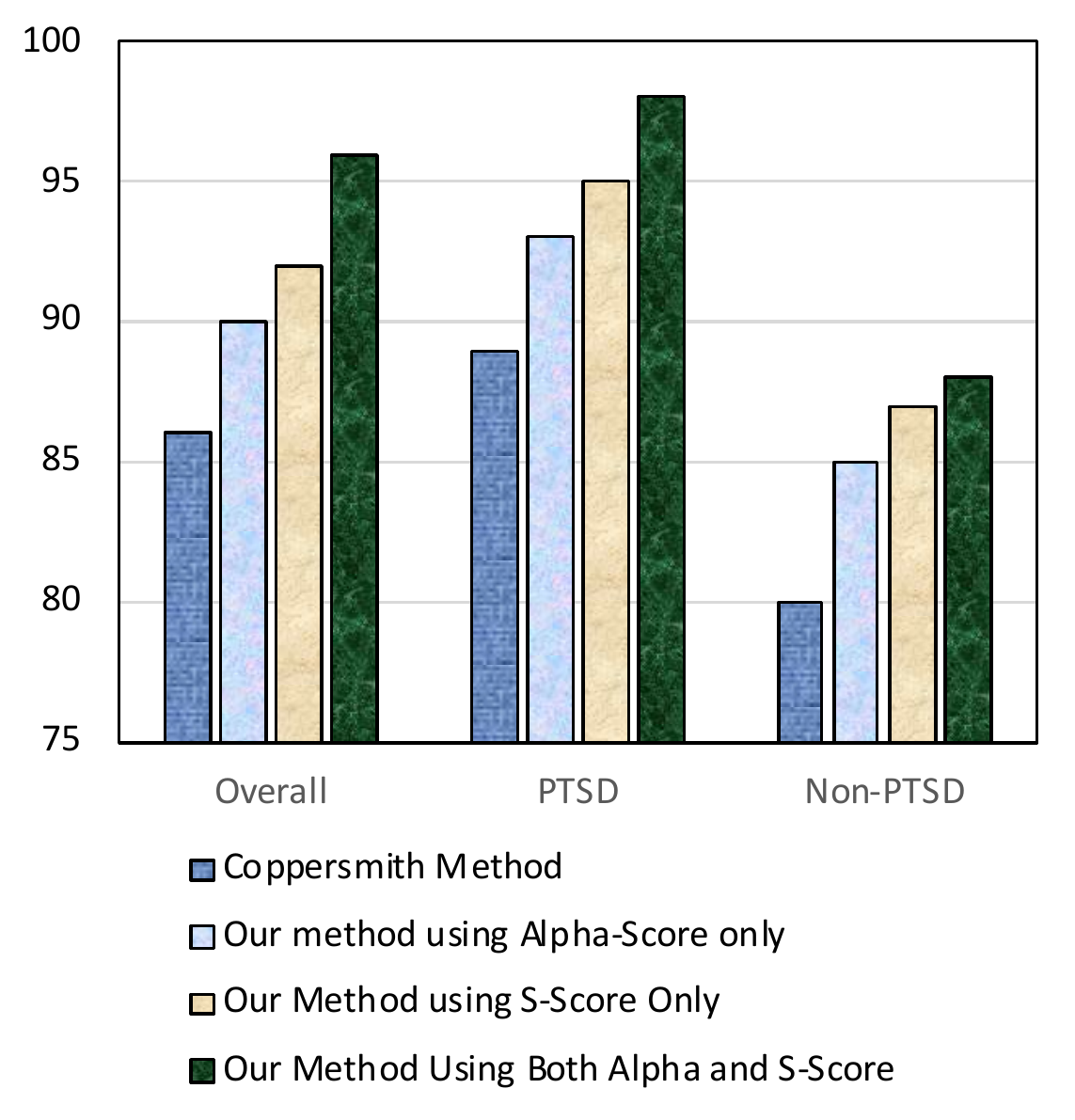,height=2.2in, width=3in}
 \caption{Comparisons between Coppersmith et. al. and our method}
 \label{fig:comparisons}
\end{center}
\end{figure}

\begin{figure}[!htb]
\begin{minipage}{0.23\textwidth}
\begin{center}
 \epsfig{file=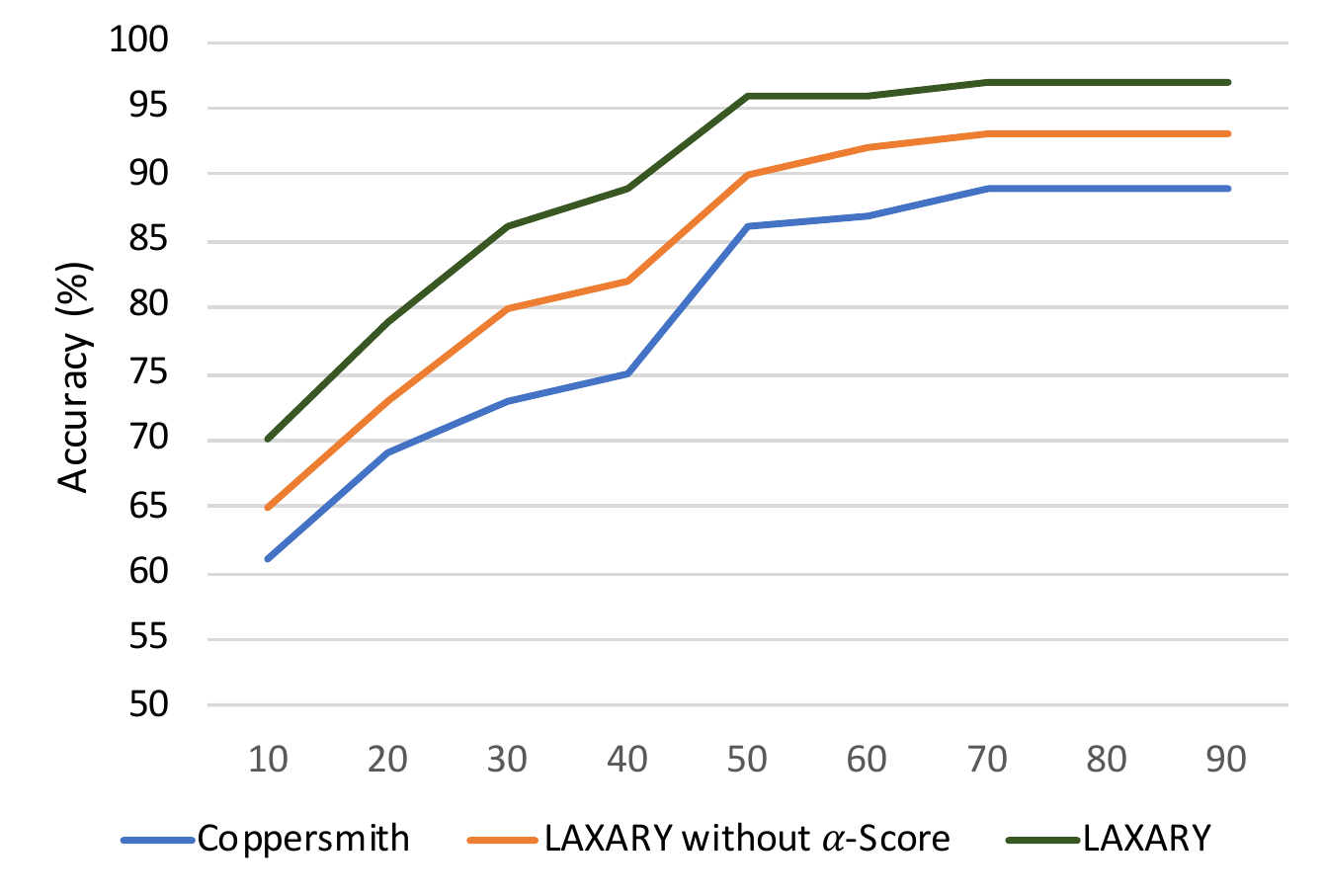,height=1.8in, width=1.8in}
 \caption{Training dataset (\%) and their PTSD detection accuracy results comparisons.  Rest of the dataset has been used for testing}
 \label{fig:compare_flow}
\end{center}
\end{minipage}
\begin{minipage}{0.23\textwidth}
\begin{center}
 \epsfig{file=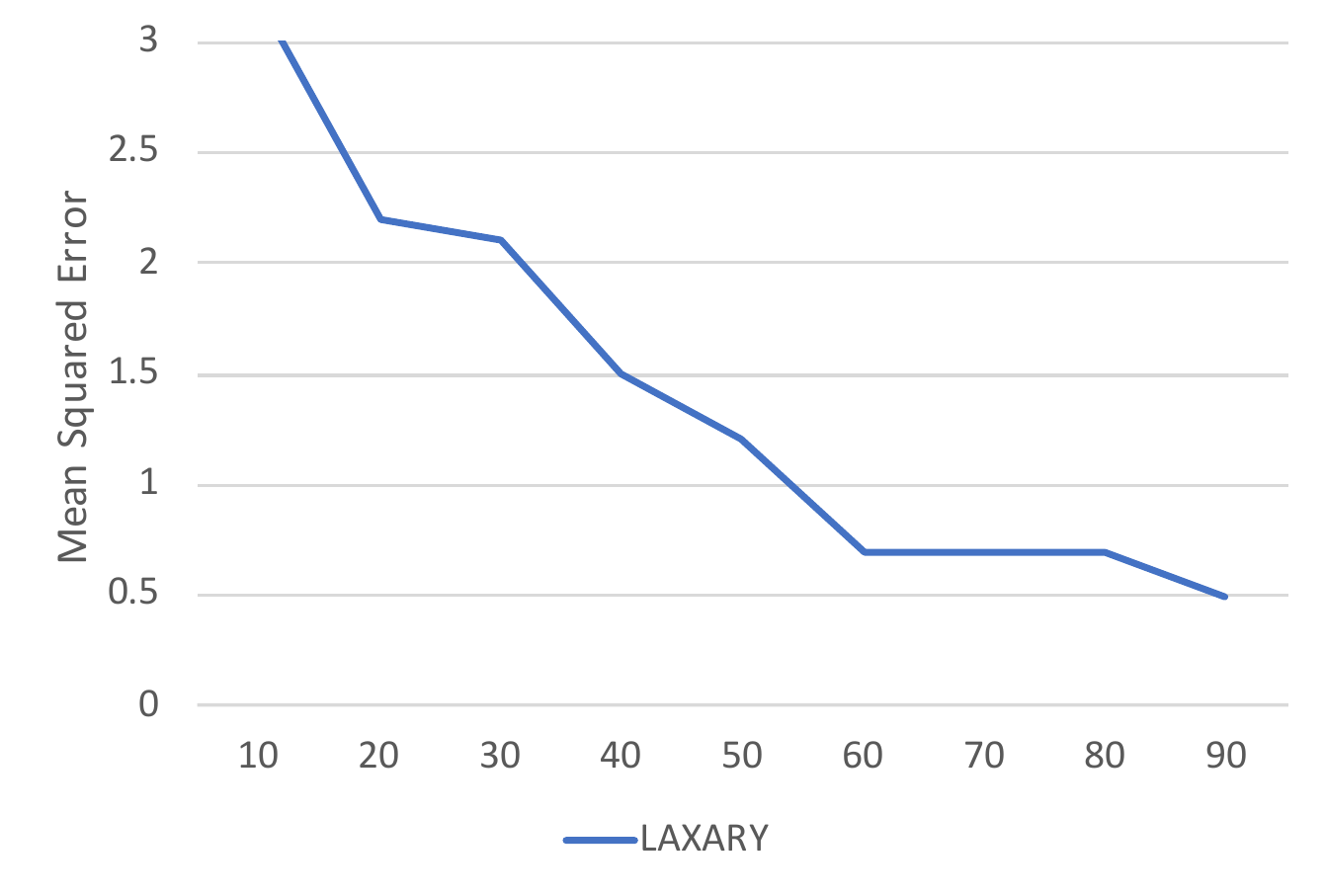,height=1.8in, width=1.8in}
 \caption{Training dataset (\%) and their Mean Squared Error (MSE) of PTSD Intensity.  Rest of the dataset has been used for testing}
 \label{fig:intensity_graph}
\end{center}
\end{minipage}
\end{figure}

\begin{figure}[!htb]
\begin{minipage}{0.23\textwidth}
\begin{center}
 \epsfig{file=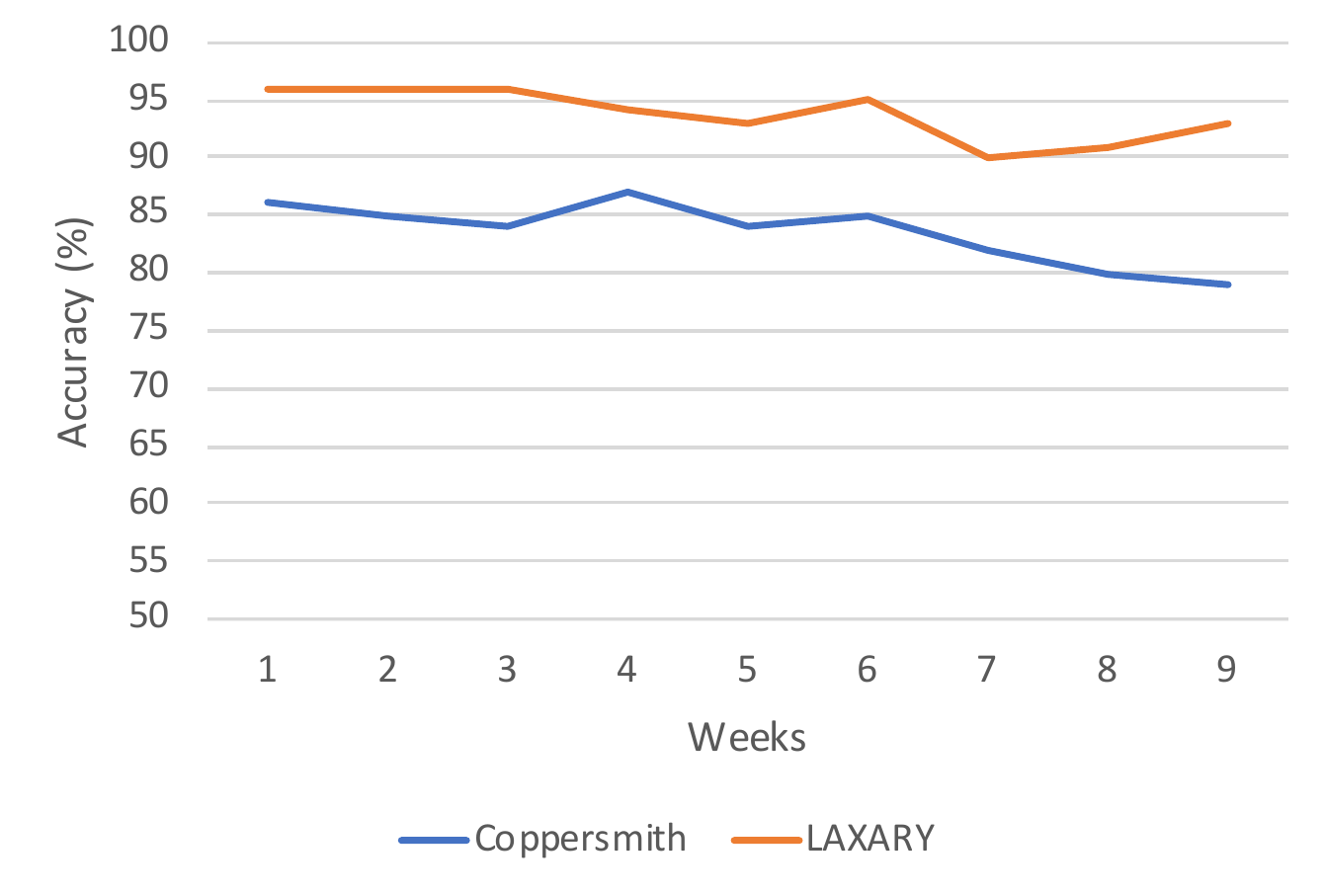,height=1.8in, width=1.8in}
 \caption{Weekly PTSD detection accuracy change comparisons with baseline model}
 \label{fig:weekly_accuracy_comparisons}
\end{center}
\end{minipage}
\begin{minipage}{0.23\textwidth}
\begin{center}
 \epsfig{file=weekly_accuracy_comparisons.pdf,height=1.8in, width=1.8in}
 \caption{Weekly PTSD detection accuracy change comparisons with baseline model}
 \label{fig:weekly_accuracy_comparisons}
\end{center}
\end{minipage}
\end{figure}

\subsection{Results}
To provide an initial results, we take 50\% of users' last week's (the week they responded of having PTSD) data to develop PTSD Linguistic dictionary and apply LAXARY framework to fill up surveys on rest of  50\% dataset. The distribution of this training-test dataset segmentation followed a 50\% distribution of PTSD and No PTSD from the original dataset. Our final survey based classification results showed an accuracy of 89\% in detecting PTSD and mean squared error of 1.8 in estimating its intensity given we have four intensity, No PTSD, Low Risk PTSD, Moderate Risk PTSD and High Risk PTSD with a score of 0, 1, 2 and 3 respectively. To compare the outperformance of our method, we also implemented Coppersmith et. al. proposed method and achieved an 86\% overall accuracy of detecting PTSD users \cite{cop14} following the same training-test dataset distribution.  Fig~\ref{fig:comparisons} illustrates the comparisons between LAXARY and Coppersmith et. al. proposed method. We also illustrates the importance of $\alpha-score$  in Fig~\ref{fig:compare_flow}. Fig~\ref{fig:compare_flow} illustrates that if we change the number of training samples (\%), LAXARY models outperforms Coppersmith et. al. proposed model under any condition. In terms of intensity, Coppersmith et. al. totally fails to provide any idea however LAXARY provides extremely accurate measures of intensity estimation for PTSD sufferers (as shown in Fig~\ref{fig:intensity_graph}) which can be explained simply providing LAXARY model filled out survey details. Fig~\ref{fig:each_survey} shows the classification accuracy changes over the training sample sizes for each survey which shows that DOSPERT scale outperform other surveys. Fig~\ref{fig:weekly_accuracy_comparisons} shows that if we take previous weeks (instead of only the week diagnosis of PTSD was taken), there are no significant patterns of PTSD detection.

\section{Conclusion}
To promote better comfort to the trauma patients, it is really important to detect Post Traumatic Stress Disorder (PTSD) sufferers in time before going out of control that may result catastrophic impacts on society, people around or even sufferers themselves. Although, psychiatrists invented several clinical diagnosis tools (i.e., surveys) by assessing symptoms, signs and impairment associated with PTSD, most of the times, the process of diagnosis happens at the severe stage of illness which may have already caused some irreversible damages of mental health of the sufferers. On the other hand, due to lack of explainability, existing twitter based methods are not trusted by the clinicians. In this paper, we proposed, LAXARY, a novel method of filling up PTSD assessment surveys using weekly twitter posts. As the clinical surveys are trusted and understandable method, we believe that this method will be able to gain trust of clinicians towards early detection of PTSD.

\bibliographystyle{SIGCHI-Reference-Format}

\end{document}